\documentclass{article}
\usepackage{spconf,amsmath,graphicx}
\usepackage{multirow,subcaption}
\usepackage{xcolor}

\title{Confidence-aware agglomeration classification and segmentation \\ of 2D microscopic food crystal images
\textsuperscript{*} 
\thanks{THIS MATERIAL IS BASED UPON WORK SUPPORTED BY THE NATIONAL SCIENCE FOUNDATION UNDER GRANT NO. 2134667.}
}
\name{Xiaoyu Ji, Ali Shakouri, Fengqing Zhu}
\address{Elmore School of Electrical and Computer Engineering, Purdue University,\\ West Lafayette, IN 47907, USA}

\begin{document}
%
\maketitle
\begin{abstract}
Food crystal agglomeration is a phenomenon occurs during crystallization which traps water between crystals and affects food product quality.
Manual annotation of agglomeration in 2D microscopic images is particularly difficult due to the transparency of water bonding and the limited perspective focusing on a single slide of the imaged sample.
To address this challenge, we first propose a supervised baseline model to generate segmentation pseudo-labels for the coarsely labeled classification dataset. Next, an instance classification model that simultaneously performs pixel-wise segmentation is trained. Both models are used in the inference stage to combine their respective strengths in classification and segmentation. To preserve crystal properties, a post processing module is designed and included to both steps. Our method improves true positive agglomeration classification accuracy and size distribution predictions compared to other existing methods. Given the variability in confidence levels of manual annotations, our proposed method is evaluated under two confidence levels and successfully classifies potential agglomerated instances.

\end{abstract}

\section{Introduction}
\label{sec:intro}

\begin{figure*}[t]
  \centering
  \includegraphics[scale=0.65]{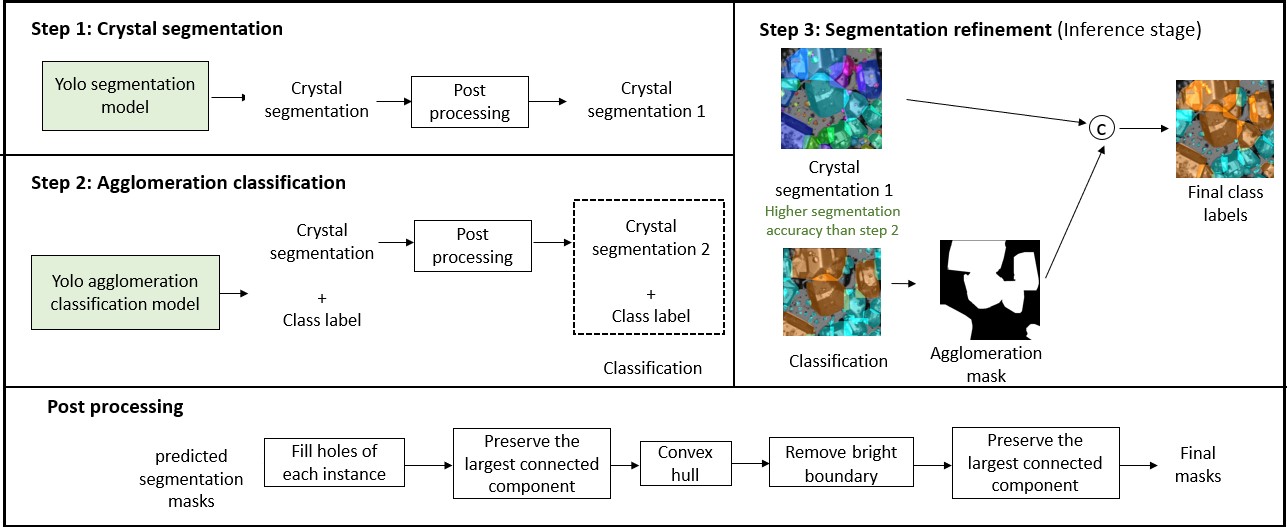}
  \caption{Overview of our proposed agglomeration classification method. The two upper left blocks represent the crystal segmentation and agglomeration classification steps. The green-shaded blocks represent the models undergoing training. The Step 1 model is utilized to generate pseudo labels for training in Step 2. The Step 2 model predicts instance segmentation and the corresponding class label. The upper right block illustrates the segmentation refinement step of the inference stage, ``c" symbol represents assigning class labels to the segmented instances based on the binary mask. Crystal segmentation result 1 is preferred to result 2 because of its higher accuracy. The ``Post processing" blocks shows the detailed steps. }
  \label{fig:1}
\end{figure*}

Crystal particle agglomeration behavior is a process where bridges form between particles, trapping water during crystallization, resulting in the clustering of large particles \cite{Alander04}. To analyze this behavior and its influence on product quality, it is crucial to measure the degree of agglomeration, defined as the percentage of single crystals that are agglomerated. Microscopic image analysis has been widely used in this scenario \cite{ho17,Timothy25,Jiang21}.

However, the transparency of food crystals and water bridges introduces ambiguity in single-view image manual annotations. There are several existing directions to address this challenge. One approach is multi-view imaging, which involves using a specially designed dual camera setup to capture images from samples \cite{ho17}. Another approach considers all overlapping crystals as pseudo-agglomerated, assigns low clarity difference pairs as true agglomerations \cite{Jiang21}. This approach is proposed for needle shape crystal analysis. A third approach involves the use of alternative techniques, such as ultrasonic irradiation, which has yet to be proven feasible \cite{Stefan19}. The last method is supervised deep learning model training proposed for sparsely distributed crystal images \cite{Timothy25}, which uses MaskRCNN \cite{He17} architecture and data augmentation strategy. However, this approach does not consider the confidence issue during manual annotation. In conclusion, the approaches discussed either employ intricate imaging setups or are tailored for sparse crystal data.

Our method relies solely on single-view microscopic images without requiring additional input. The crystals are densely distributed, making it difficult to distinguish agglomerations, particularly when one crystal is obscured by another. We invite domain experts to annotate only the agglomerations with highly confidence. Our model then predicts potential agglomerations and aims at classifying all high confidence annotations correctly.

Similar to existing food crystal analysis works \cite{Timothy25,Jiang21,ji24}, our method is based on training deep learning models. MaskRCNN \cite{He17}, Yolo \cite{yolov8} and Stardist \cite{schmidt18} are widely used instance segmentation architectures for microscopic images \cite{Jo21,ji24,Weigert22}. Mask-scoring R-CNN \cite{huang19} is an extended work of MaskRCNN which calibrates the predicted instance mask confidence scores. Except for the Stardist model, which has limited capability in overlapping instances predictions, all the other models are used as the baseline models in this paper.

Our method includes three steps. Firstly, a supervised instance segmentation model is trained for crystal segmentation. This model is used to generate pseudo segmentation annotations for coarsely labeled datasets, reducing the need for manual effort. In the second step, we train an agglomeration classification model, which is a segmentation and classification two-branch model predicting agglomeration classification label for each segmented instance. The last step is the inference stage, which integrates the segmentation outputs of the first step and the classification outputs of the second step to improve segmentation accuracy.

In the evaluation process, we test segmentation accuracy at two confidence levels. We prepare the segmentation dataset with manual annotated confidence levels and the evaluation is conducted separately for each confidence level. The classification dataset with ground truth agglomeration class masks is used to measure the true positive rate of classification.

\section{Method}
In this section, we first describe the two training steps including the crystal segmentation model and the agglomeration classification model. Then we introduce the refinement implemented in the inferencing process. The post processing segmentation refining module added to segmentation predictions is finally introduced.

\subsection{Crystal segmentation model training}
\label{sec:method1}
The Yolo segmentation model shown in the upper left step 1 block of Fig. \ref{fig:1} is trained using Yolov8 instance segmentation architecture \cite{yolov8}. The backbone network architecture consists of five layers, with the segmentation head utilizing the output features from the top three layers of the backbone. It is supervisedly trained with fully labeled data and partially labeled data which will be discussed in Section \ref{sec:data}.

\subsection{Agglomeration classification model training}
The classification segmentation model is illustrated in step 2 of Figure \ref{fig:1}. This model predicts instance agglomeration class and pixel-wise segmentation. The training dataset used segmentation model in step 1 to generate the pseudo segmentation labels. This model is trained using the same backbone Yolov8 \cite{yolov8} architecture as step 1, with an additional classification head. The loss function is shown in (\ref{eq:1}). 

\begin{equation}
\centering
\begin{aligned}
\label{eq:1}
loss=&\alpha_1loss_{segm}+\alpha_2loss_{bbox}+\\
&\alpha_3loss_{cls}+\alpha_4loss_{DFL}\\
\end{aligned}
\end{equation}

In Equation (\ref{eq:1}), the loss function includes the bounding box error ($loss_{bbox}$), segmentation loss ($loss_{segm}$), classification loss ($loss_{cls}$) and Distribution Focused Loss (DFL) ($loss_{DFL}$). The detailed definitions can be found in \cite{yolov8}. The loss weights of the classification model $\alpha_1,\alpha_2,\alpha_4$ are different from the settings of the segmentation model to balance the pre-trained knowledge with the integration of new classification information. The fine-tuned configuration settings are listed in Section \ref{sec:exp1}.

\subsection{Segmentation refinement}
The segmentation refinement process is implemented during the inference stage. We observe a trade-off between classification accuracy and segmentation accuracy during agglomeration classification model training. We achieve a good classification accuracy but a lower segmentation accuracy than the segmentation model in Section \ref{sec:method1}. Therefore, we combine the results of the two models: predicted instances of the segmentation model with over 50\% overlap with the predicted agglomeration region from the classification model are classified as agglomeration. This approach enables us to achieve better performance in both segmentation and classification predictions simultaneously.

\subsection{Post processing}
The post processing pipeline is shown in the bottom part of Fig. \ref{fig:1}. It consists of five morphological operations and the figure shows an example of each corresponding processed result. The first two operations involve infilling holes in the predicted mask and preserving the largest connected component to ensure the enclosing property of the crystals. Next, we emphasize the convex property of the crystals by applying a convex hull operation to the predicted crystals. Since this operation can extend the boundary, we remove the bright boundary within the top 15\% of the brightest colors, as the crystal edges are typically darker than the background. Finally, in rare cases where boundary removal splits the mask into multiple components, we repeat the step of preserving only the largest connected component.

\section{Dataset}
\label{sec:data}

\subsection{Segmentation dataset}
Our dataset consists of microscopic images of food solution samples taken during an intermediate stage of the manufacturing process. The liquid product was placed between two glass slides for imaging. 

There are two segmentation datasets we used in this study. \textit{Dataset 1} is a fully pixel-wisely labeled dataset comprising 24 training images, 7 validation images and 12 test images, each sized 512$\times$512. All the annotations were completed by one operator, encompassing high and low confidence levels. \textit{Dataset 2} is a partially labeled dataset with 156 training images and 48 validation images. In this dataset, only high-confidence crystals are labeled by a domain expert different from the annotator of \textit{Dataset 1}. To avoid the model from misclassifying unlabeled crystals as background, a blur filter with a window size of $33\times 33$ pixels is applied (example shown in Figure \ref{fig:1}). The annotated confidence levels are only used in the evaluation process.

\subsection{Classification dataset}
Due to the difficulty in distinguishing agglomerated crystals, we consulted a domain expert to manually circle the regions they consider as agglomerated (example shown in Figure \ref{fig:1}). There are 63 images  that are coarsely labeled, we name this dataset as \textit{Dataset 3}. We utilize the pre-trained segmentation model to generate instance masks. Instances with an overlap ratio exceeding 50\% with the agglomerated region are labeled agglomeration. This dataset is split into 43 training images, 11 validation images and 9 test images.

  \section{Experiments}

\subsection{Experimental setting}
\label{sec:exp1}

Segmentation models are trained using the training images from \textit{Dataset 1} and \textit{Dataset 2} (180 training images and 55 validation images), given ground truth instance masks and confidence labels. The training and testing processes are performed on one NVIDIA GeForce GTX 1080 Ti GPU. 

We initialize Yolov8 model using the pre-trained weights from the COCO dataset \cite{coco14} and train it for a maximum of 1,000 epochs, employing early stopping with a tolerance of 100 epochs. We use the Adam optimizer \cite{adam14} with $\beta_1 = 0.937$, $\beta_2 = 0.999$ and $w=5\times10^{-4}$. The learning rate is initialized as 0.01. The loss weight $\alpha_1$ and $\alpha_2$ are set at 7.5, $\alpha_3$ at 0.5 and $\alpha_4$ at 1.5. 

Classification models are trained with \textit{Dataset 3}. The parameter settings of Yolov8 model are the same as those of the segmentation model, except that the segmentation and bounding box loss weights ($\alpha_1$ and $\alpha_2$ in Equation (\ref{eq:1})) are fine-tuned to 9, the DFL loss weight $\alpha_4$ is increased to 3. This adjustment enhances the feature learning for the minor class (agglomeration class) segmentation. This model is initialized using the pre-trained COCO segmentation model weights. The parameter settings of the compared Mask RCNN model are also fine-tuned.

\subsection{Evaluation at two confidence levels}
\begin{figure}[t]
  \centering
  \includegraphics[scale=0.35]{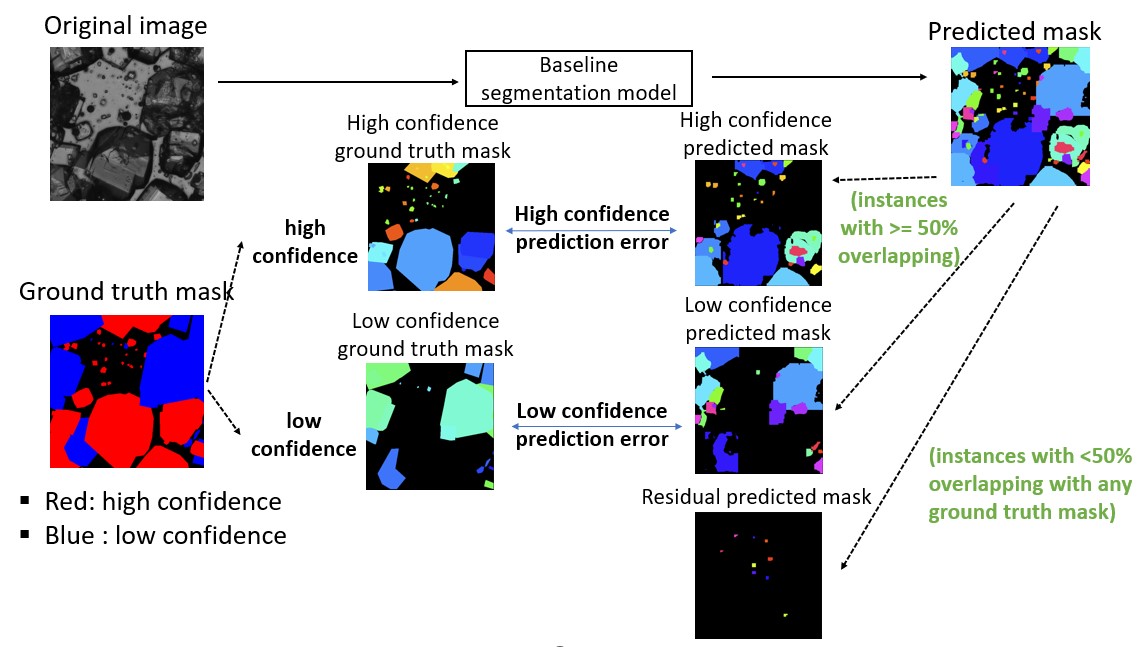}
  \caption{Overview of two confidence level evaluation process. The predicted mask on the right is split into high confidence, low confidence and residual masks, paired with the high and low confidence ground truth masks. }
  \label{fig:2}
\end{figure}

The evaluation process of the segmentation predictions takes two confidence levels into consideration, as shown in Figure \ref{fig:2}. The two images on the left are the original input image and the ground truth mask, where red represents high confidence and blue represents low confidence. The upper right image is the predicted mask from the segmentation model. For each predicted instance, the overlap ratios with both high-confidence and low-confidence ground truth masks are measured. If the largest overlapping ratio is less than 50\%, the predicted instance is included in the residual mask. Otherwise, it is assigned to the confidence level with the higher ratio. The prediction error metrics are measured separately for the predicted masks at the two confidence levels and the residual mask.

\subsection{Segmentation model results}
\label{sec:exp}

\begin{table*}[htbp]
\caption{Results of baseline segmentation models w/ or w/o post processing}
\centering  
\label{tab:table_1}
\normalsize
\begin{tabular}{|p{0.25\columnwidth}|p{0.17\columnwidth}|p{0.18\columnwidth}|p{0.18\columnwidth}|p{0.18\columnwidth}|p{0.18\columnwidth}|p{0.19\columnwidth}|p{0.2\columnwidth}|p{0.1\columnwidth}|} 
\hline
Models & post \par processing& high conf corr (\%) $\uparrow$ & high conf $\chi^2$ (\%) $\downarrow$ & low conf corr (\%) $\uparrow$ & low conf $\chi^2$ (\%) $\downarrow$ & high conf mAP (\%) $\uparrow$ & high conf recall (\%) $\uparrow$ & ResErr (\%) $\downarrow$\\
\hline
\hline
\multirow{2}{*}{M-RCNN \cite{He17}} & w/o & 45.452 & 73.615 & 15.983 & 41.410 & 58.162 & 42.501 & 9.590\\
\cline{2-9}
 & w/ & \underline{46.275} & \underline{68.686} & \underline{21.745} & \underline{39.159} & \underline{60.178} & \underline{43.730} & \underline{9.297}\\
\hline
\multirow{2}{*}{Ms-RCNN \cite{huang19}} & w/o & 42.521 & 73.978&17.906&24.749&\underline{54.817}&\underline{42.908} & 10.301\\
\cline{2-9}
& w/& \underline{48.932}&\underline{69.189}&\underline{23.768}&\underline{21.717}&53.183&41.862 & \underline{9.523}\\
\hline
\multirow{2}{*}{Yolov8 \cite{yolov8}}&  w/o & 96.471&9.115&70.683&\underline{\textbf{16.459}}&72.139&65.976 & 7.920\\
\cline{2-9}
& w/ & \underline{\textbf{97.118}}&\underline{\textbf{8.139}}&\underline{\textbf{74.650}}&16.666&\underline{\textbf{73.056}}&\underline{\textbf{67.453}} & \underline{\textbf{7.262}}\\
\hline
\end{tabular}
\end{table*}

As shown in Table \ref{tab:table_1}, baseline models \cite{He17,huang19,yolov8} with or without post processing have been compared with seven evaluation metrics. The four metrics on the left are the crystal size distribution measurements. Metrics measure between high-confidence ground truth and prediction masks are indicated as ``high conf", low-confidence metrics are indicated as ``low conf". ``corr" represents the ground truth and predicted crystal size distribution correlation, and $\chi^2$ represents the mean squared crystal size error between each bin of the ground truth and predicted size distributions (with a total of 10 bins). The last three metrics are instance-wise segmentation accuracy metrics, Mean Average Precision (mAP) and recall using an intersection-over-union (IoU) confidence threshold of 50\%. mAP indicates the ratio of predicted crystals that are accurate while recall indicates the ratio of ground truth instances that are predicted. ``ResErr" represents the ratio of predicted crystals in the residual mask, which is the instance-wise false positive rate.

The bold numbers in Table \ref{tab:table_1} are the best value for each metric. The underscored values indicate the better result for each model when comparing performance with and without post processing. Segmentation models with post processing have better performances than those without for most of the metrics, especially for high-confidence metrics. Because low-confidence ground truth indicates uncertainty inherent in manual annotation process, a lower accuracy on low-confidence metrics is observed than those on high-confidence metrics as expected.  

In general, the Yolov8 model outperforms other baseline models in all the evaluation metrics. The post processing module incorporates intrinsic properties of food crystals which are not learned in the models, improving size distribution predictions and instance-wise prediction accuracy. Yolov8 model with post processing is selected to be the pseudo-label generation model.

\subsection{Classification model results}

The final inference stage predicts both the instance segmentation mask and the corresponding class label. Classification results are evaluated on the 9 test images in \textit{Dataset 3}. Because only high-confidence agglomerated crystals are annotated in the ground truth, we only evaluate the ratio of the ground truth agglomeration area that overlaps with correctly classified predicted masks, which is the True Positive rate (TPR) shown in Table \ref{tab:table_3}. The other two metrics are the high-confidence size distribution metrics for segmentation evaluation at the same time.

As shown in Table \ref{tab:table_3}, we reconstruct the agglomeration classification methods \cite{Timothy25}, the results are shown in the first row. Yolov8 method represents the direct results of the model in Step 2. We can observe that the second and third size distribution metrics have a poorer performance compared to the results of the last row of Table \ref{tab:table_1}. Therefore, we combine the segmentation results of the step 1 segmentation model and the classification results of the step 2 model to get the final results which is the last row in the table.

As expected, our method achieves a better size distribution correlation and $\chi^2$ values than other methods. It is exciting to observe a higher classification accuracy for our method, which is 91.079\% than the Yolov8 model. This observation indicates that the predicted masks of the segmentation model cover a larger portion of the ground truth agglomeration mask than the classification model. Although accuracy is increased, a limitation of our method is the increased time complexity during inference due to the use of two models.

\begin{table}[htbp]
\caption{Agglomeration classification and segmentation metrics. Both the ``corr" and $\chi^2$ metrics are in the high confidence level.}
\centering  
\label{tab:table_3}
\normalsize
\begin{tabular}{|p{0.35\columnwidth}|p{0.1\columnwidth}|p{0.18\columnwidth}|p{0.18\columnwidth}|} 
\hline
Methods & TPR (\%)$\uparrow$&  high conf corr (\%) $\uparrow$& high conf $\chi^2$ (\%) $\downarrow$\\
\hline
M-RCNN \cite{He17} & 77.109 & 58.085 & 143.135\\
\hline
Yolov8 \cite{yolov8} &  90.285 & 94.272 & 17.035\\
\hline
Ours &  \textbf{91.079} & \textbf{97.118} & \textbf{8.139}\\
\hline
\end{tabular}
\end{table}

\subsection{Qualitative Results}

\begin{figure}[!ht]
  \centering
   \subcaptionbox
      {Input image\label{result:a}}{\includegraphics[scale=0.5]{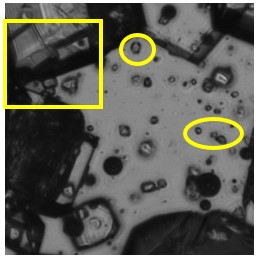}}
      \hspace{1ex}
    \subcaptionbox
      {Ground truth\label{result:b}}{\includegraphics[scale=0.5]{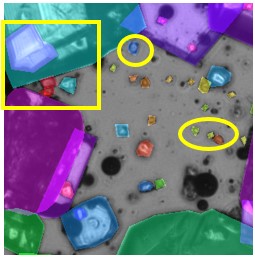}}%
      \par
    \subcaptionbox
      {Mask RCNN \cite{He17} w/ post\label{result:f}}{\includegraphics[scale=0.42]{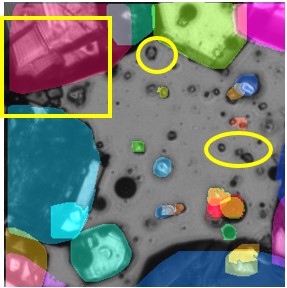}}%
     \hspace{2ex}
    \subcaptionbox
      {Mask-scoring RCNN \cite{huang19} w/ post \label{result:g}}{\includegraphics[scale=0.42]{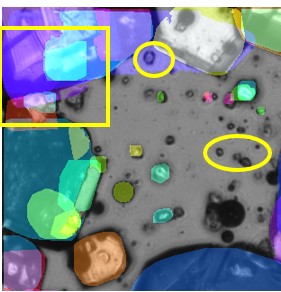}}%
      \hspace{2ex}
    \subcaptionbox
      {Yolov8 \cite{yolov8} w/ post\label{result:h}}{\includegraphics[scale=0.38]{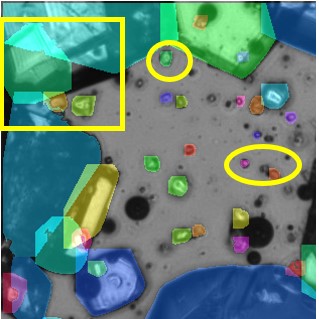} }
  \caption{Visualization comparisons of different baseline segmentation methods with post processing on food crystal dataset. Subfigure (a)-(e) are the input image, the ground truth label image, results of Mask RCNN \cite{He17}, Mask-scoring RCNN \cite{huang19} and Yolov8 \cite{yolov8} with post processing. Highlighted in yellow are the key areas where model performance variations are most noticeable.}
  \label{fig:3}
\end{figure}

In this section, we first visually compare the segmentation results of different baseline methods, as shown in Fig. \ref{fig:3}. Fig. \ref{result:a} and Fig. \ref{result:b} are the original image and the ground truth labeled image. Each overlaid colored mask in Fig. \ref{result:b} represents a crystal instance. Fig. \ref{result:f}, Fig. \ref{result:g} and Fig. \ref{result:h} are the prediction results of Mask RCNN \cite{He17}, Mask-scoring RCNN \cite{huang19} and Yolov8 \cite{yolov8} segmentation models with post processing.

There are three regions of interest highlighted to compare the three models with post processing. The yellow rectangle region contains largely overlapping crystals, where the Yolov8 prediction results in Fig. \ref{result:h} show more accurately predicted instances compared to those of other models. The boundary of the large crystal does not align well due to the uncertainty in manual annotations of clustered crystals in the training dataset. The two yellow circled regions focus on small crystals, where all of them are missing or incorrectly segmented in Fig. \ref{result:f} and Fig. \ref{result:g}. Yolov8 model correctly segmented more small crystals than other models, which corresponds to a high size distribution accuracy. To conclude, Yolov8 model with post processing has a higher segmentation accuracy than other methods.

\begin{figure}[!ht]
  \centering
   \subcaptionbox
      {Input image\label{result:4a}}{\includegraphics[scale=0.13]{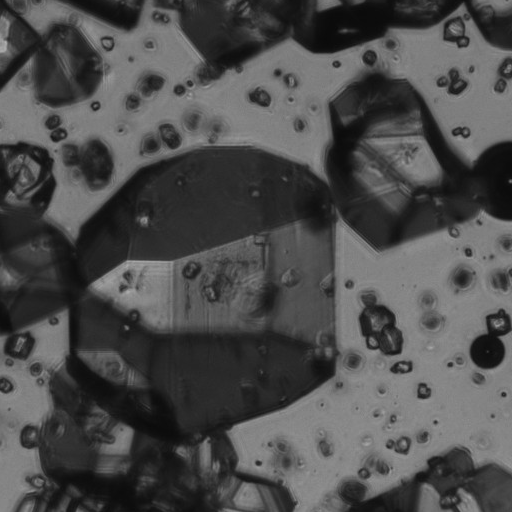}}%
      \hspace{2ex}
    \subcaptionbox
      {Ground truth\label{result:4b}}{\includegraphics[scale=0.13]{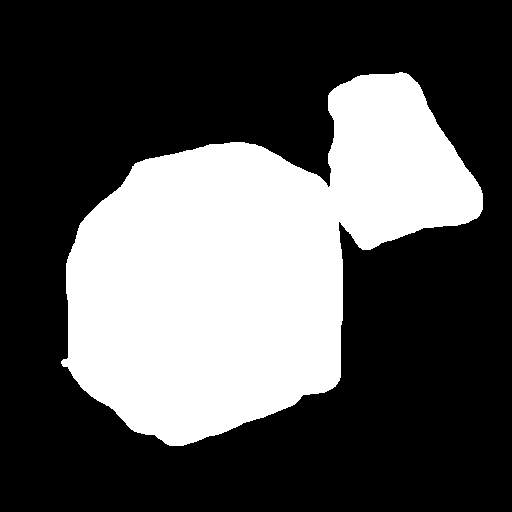}}%
      \par
      \hspace{2ex}
    \subcaptionbox
      {M-RCNN \cite{He17} \label{result:4c}}{\includegraphics[scale=0.13]{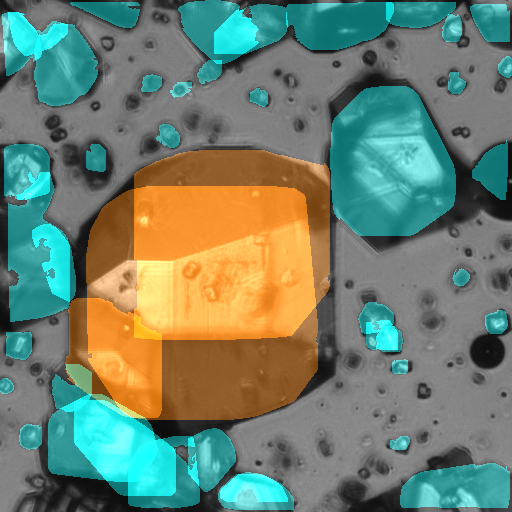}}%
       \hspace{2ex}
      \subcaptionbox
      {Yolov8 \cite{yolov8} \label{result:4d}}{\includegraphics[scale=0.13]{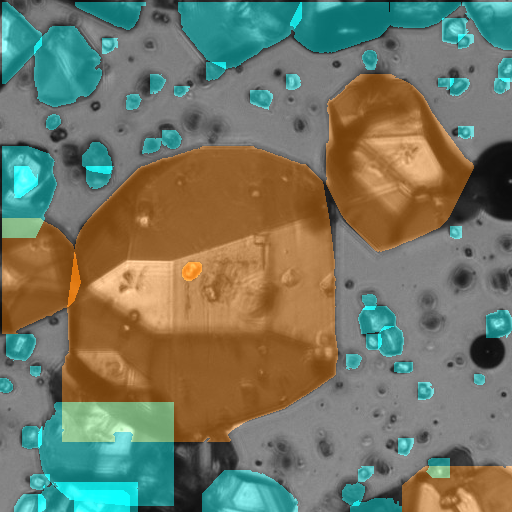}}%
       \hspace{2ex}
      \subcaptionbox
      {Ours \label{result:4e}}{\includegraphics[scale=0.13]{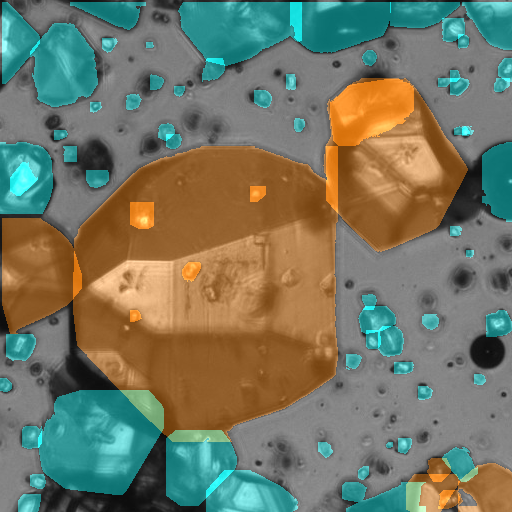}}%
      \par

  \caption{Visualization comparisons of agglomeration classification. Subfigures (a)-(e) are the input image, ground truth high-confidence agglomeration mask, the results of compared methods \cite{Timothy25,yolov8} and our method.}
  \label{fig:4}
\end{figure}

The visual result of classification is shown in Fig. \ref{fig:4}. The ground truth mask in Fig. \ref{result:4b} shows the agglomeration areas labeled with high confidence. Note that crystals not classified as agglomerated in the ground truth may still have the potential to be agglomerated. There are two crystals classified as high confident agglomerations in this example. One crystal is missing in the result of Mask RCNN with data augmentation strategy method \cite{Timothy25}, while both are correctly classified in the Yolov8 result image (Fig. \ref{result:4d}). Yolov8 model also predicts other crystals that have the potential to be agglomerated. Our method compared to Yolov8 model has a better segmentation result, we can observe the small crystals on the large agglomerated crystal which provides more details.

\section{Conclusion}
In this paper, we propose a method for agglomeration classification and segmentation of food crystal microscopic images. Our method involves a two-step training process and integrates the results of both models in the inference stage to take their strengths in segmentation and classification simultaneously. We design a post processing module to preserve crystal properties of the predictions. Given the ambiguity in manual annotation, confidence levels are assigned and incorporated into the evaluation process.

\bibliographystyle{IEEEbib}
\bibliography{main}

\end{document}